\documentclass[runningheads]{llncs}
\usepackage{graphicx}
\usepackage{floatrow}
\newfloatcommand{capbtabbox}{table}[][\FBwidth]
\begin{document}
\title{A Filtering-based General Approach to Learning Rational Constraints of Epistemic Graphs}
\titlerunning{A Filtering-based General Approach to Learning Constraints}
%
\author{Xiao Chi
}
\authorrunning{X. Chi}
%
\institute{Guanghua Law School, Zhejiang University, Hangzhou 310008, China 
}
%
\maketitle              
\begin{abstract}
Epistemic graphs are a generalization of the epistemic approach to probabilistic argumentation. 
 Hunter proposed a 2-way generalization framework to learn epistemic constraints from crowd-sourcing data. However, the learnt epistemic constraints only reflect users' beliefs from data, without considering the rationality encoded in epistemic graphs. Meanwhile, the current framework can only generate epistemic constraints that reflect whether an agent believes an argument, but not the degree to which it believes in it. The major challenge to achieving this effect is that the computational complexity will increase sharply when expanding the variety of constraints, which may lead to unacceptable time performance. To address these problems, we propose a filtering-based approach using a multiple-way generalization step to generate a set of rational rules which are consistent with their epistemic graphs from a dataset. This approach is able to learn a wider variety of rational rules that reflect information in both the domain model and the user model. Moreover, to improve computational efficiency, we introduce a new function to exclude meaningless rules. The empirical results show that our approach significantly outperforms the existing framework when expanding the variety of rules.
 

\end{abstract}
\section{Introduction}
Argumentation plays an important role in daily life when there exist conflicts containing inconsistent and incomplete information \cite{DBLP:journals/ijar/Hunter13}. Usually, argumentation is often pervaded with uncertainty either within an argument or between arguments. Some research efforts have been paid attention to dealing with various kinds of uncertainties of arguments. Among others, a feasible solution is to quantify the uncertainty of arguments by assigning probability values to arguments. These probability values are used to represent the degrees of beliefs of agents towards arguments \cite{DBLP:conf/comma/Hunter12,DBLP:conf/tafa/LiON11,DBLP:conf/ecai/Thimm12,DBLP:conf/comma/DungT10}. A typical problem of fundamental probabilistic argumentation is that it only considers attack relations. This problem can be resolved by using epistemic graphs which can model support relation, attack relation and neither support nor attack relation. Furthermore, there has been another progress on epistemic graphs by restricting beliefs one has in arguments, and indicating how beliefs in arguments influence each other with varying degrees of specificity using epistemic constraints \cite{DBLP:journals/ai/HunterPT20}. It has been evidenced that epistemic graphs can be applied to an automated persuasion system (APS) to represent domain models, and probability distributions over arguments can be used to represent user models \cite{DBLP:journals/ai/HunterPT20}. The APS persuades a user to believe (or disbelieve) something by offering arguments that have a high probability of influencing the user \cite{hunter2016computational}. The domain model together with the user model is harnessed by the strategy of APS for choosing good moves in a persuasion dialogue. When selecting moves in persuasion dialogues, a choice needs to be made between relying on either the domain model or the user model. In this process, the user model is represented in terms of epistemic constraints. In order to learn a set of constraints that can be harnessed directly by the strategy of APS, the following two aspects need to be considered.

On the one hand, to obtain such constraints, Hunter proposed a framework to learn them from crowd-sourcing data \cite{DBLP:conf/comma/Hunter20}. These constraints only reflect the information of the user model, and may bring about the following problem: The constraints cannot reflect the rationality encoded in an epistemic graph, and therefore cannot be directly used by the APS strategy. As a result, the APS strategy has to decide whether to rely on the domain model or the user model when making choices of moves in a persuasion dialogue. Given that the domain model is not utilized in learning epistemic constraints, the learnt epistemic constraints might be irrational.

To describe the above ideas, consider the following example. In this example, an epistemic graph is used to present a domain model, and a table containing crowd-sourcing data is used to represent a user model, i.e., beliefs of agents towards the arguments in the graph. We say that an agent believes (resp. disbelieves) an argument to some degree if the degree of its belief is greater than $0.5$ (resp. less than $0.5$). 

\begin{figure}
\centering
\includegraphics[width=\textwidth]{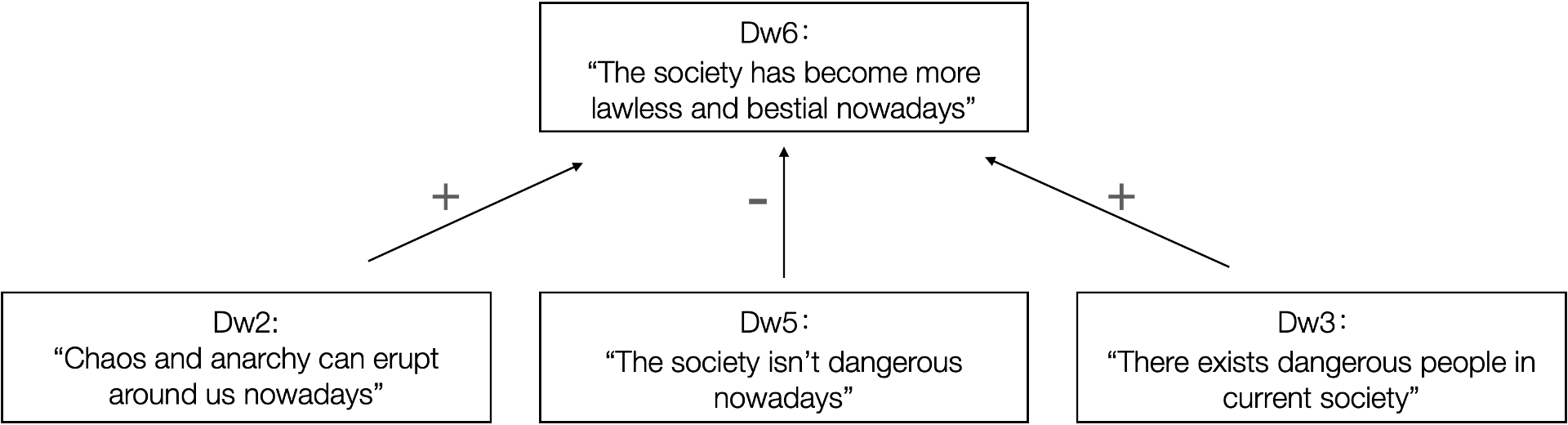}
\caption{An epistemic graph representing a domain model. 
Here, ``$+$" denotes support relation while ``$-$" denotes attack relation. } \label{fig:ex1}.
\end{figure}

\begin{example}
As illustrated in Fig. \ref{fig:ex1} and Table \ref{tab:ex1}, assume that one wants to persuade an agent to believe argument \textit{Dw6}, which is influenced by \textit{Dw2}, \textit{Dw3}, and \textit{Dw5}. Consider row $026$ of Table \ref{tab:ex1}, one may interpret the data as a constraint as follows: if \textit{Dw5} is disbelieved, and \textit{Dw2} and \textit{Dw3} are believed, then \textit{Dw6} is disbelieved. This can be represented as: 
\begin{math}
  p(\textit{Dw2})> 0.5 \wedge p(\textit{Dw5}) < 0.5 \wedge p(\textit{Dw3}) > 0.5 \rightarrow p(\textit{Dw6}) < 0.5.
\end{math}
This is how the framework proposed by Hunter \cite{DBLP:conf/comma/Hunter20} generates constraints. However, this result is inconsistent with the rationality encoded in the epistemic graph. Specifically, in terms of the attack and support relations contained in the graph, it is reasonable to infer that: if an agent disbelieves $\textit{Dw5}$, and believes $\textit{Dw2}$ and $\textit{Dw3}$, then he tends to believe $\textit{Dw6}$. 
\label{example1}
\end{example}

On the other hand, in order to persuade users, epistemic constraints need to better reflect the beliefs of participants in arguments. However, the current framework can only generate epistemic constraints that reflect whether an agent believes an argument, but not the degree to which it believes in it. The major challenge to achieving this effect is that the complexity of generating epistemic constraints will increase sharply when expanding the variety of constraints, which may lead to unacceptable time performance.


\begin{table}
\caption{Containing columns and rows of data obtained from the Italian study \cite{PELLEGRINI2019104144}.}\label{tab:ex1}
\centering
\resizebox{45mm}{8mm}{
    \begin{tabular}{|c|c|c|c|c|}
    \hline
    & Dw6 & Dw2 & Dw5 & Dw3    \\
    \hline
    004 & 0.2 & 0.3 & 0.3 & 0.3\\
    026 & 0.4 & 0.6 & 0.3 & 0.6\\
    111 & 0.6 & 0.1 & 0.6 & 0.2\\
    \hline
    \end{tabular}
}

\end{table}

The above analysis gives rise to the following three research questions. 
\begin{itemize}
    \item[1.] How to develop a method to generate constraints that can reflect information in both the domain model and the user model?
    \item[2.] How to make the method more general such that various degrees of beliefs can be represented?
    \item[3.] How to improve the efficiency of computation when expanding the variety of rules?
\end{itemize}

To address the first research question, we propose a novel method to generate a set of epistemic constraints that can be directly harnessed by the strategy of APS, i.e., the set of constraints that can reflect information in both epistemic graphs and crowd-sourcing data. Note that the learnt constraints are in the form of rules, and therefore in this paper, epistemic constraints and rules have the same meaning. Since the rules can reflect the rationality encoded in the epistemic graphs, we call this kind of rules rational rules. More specifically, this method is realized by adding a set of rationality principles as measurements to filter out irrational rules, so that the resulting set of rules is based on both the domain model and the user model. Second, we propose a multiple-way generalization step that builds upon a 2-way generalization step introduced by Hunter \cite{DBLP:conf/comma/Hunter20}, allowing us to learn a wider variety of rules. Third, we put forward a \textsf{Nearest} function to be used in the generalization step so that we are able to generate a wider variety of rules with acceptable time performance. Additionally, we evaluate our approach on datasets from two published studies, i.e., the appropriateness of using Wikipedia in teaching in Higher Education \cite{asi.23488} and political attitudes in Italy \cite{PELLEGRINI2019104144}. We compare the results of our approach with those of the framework using the 2-way generalization step.

This paper is organized as follows. Section 2 reviews the preliminary knowledge required for this paper. Section 3 introduces a framework using a 2-way generalization step. Section 4 introduces a filtering-based approach using a multiple-way generalization step by providing new definitions. Section 5 evaluates the improved approach on two datasets by comparing it with the ordinary framework. Section 6 discusses the contributions, limitations, related work and future work of this paper.

\section{Preliminaries}
This section introduces the notions of epistemic graphs and restricted epistemic languages, and sets forth background knowledge for formulating a set of new rationality principles.

An epistemic graph is a labelled graph equipped with a set of epistemic constraints, which are represented by the epistemic language. To simplify the presentations and evaluations, we use a restricted language instead of full power of the epistemic language in this paper \cite{DBLP:journals/ai/HunterPT20}. In the restricted epistemic language, probability functions take on a fixed and finite set of values. We start by introducing the definition of restricted value sets. Note that restricted value sets are closed under addition and subtraction.

\begin{definition}
Let $\Pi$ be a unit interval and $1 \in \Pi$. A finite set of rational numbers from $\Pi$ is a restricted value set iff for every $x,y \in \Pi$, it satisfies the following constraints: 1) If $x+y \leq 1$, then $x + y \in \Pi$. 2) If $x-y \geq 0$, then $x-y \in \Pi$.
\end{definition}

Now, let us formally introduce the restricted epistemic language based on a restricted value set \cite{DBLP:conf/comma/Hunter20}. 
\begin{definition}
Let $\Pi$ be a restricted value set, and $\mathrm{Nodes}(G)$ be a set containing arguments in a directed graph $G$. The restricted epistemic language based on $G$ and $\Pi$ is defined as follows:
\begin{itemize}
    \item An \textbf{epistemic atom} is defined as $p(\alpha) \# x$, where $\# \in \{=,\neq,\geq,\leq,>,<\}$, $x \in \Pi$ and $\alpha \in \mathrm{Nodes}(G)$. 
    \item An \textbf{epistemic formula} is defined as a Boolean combination of epistemic atoms.
\end{itemize}
\end{definition}

Having defined the syntax of the restricted epistemic language, we now move on to formulating its semantics, which is represented by belief distributions. A belief distribution on $\mathrm{Nodes}(G)$ is a probability distribution $P: 2^{\mathrm{Nodes}(G)} \rightarrow [0,1]$ such that $ \sum_{\Gamma \subseteq \mathrm{Nodes}(G)} P(\Gamma) = 1$. When $P$ is a belief distribution for a restricted value set $\Pi$, for every $\Gamma \subseteq \mathrm{Nodes}(G)$, $P(\Gamma) \in \Pi$.

We regard each $\Gamma \subseteq \mathrm{Nodes}(G)$ as a possible world. The probability of an argument is defined as the sum of the probabilities of its possible worlds: $P(\alpha) = \sum_{\Gamma \subseteq \mathrm{Nodes}(G)\ s.t.\ \alpha \in \Gamma}P(\Gamma)$. According to \cite{DBLP:journals/ai/HunterPT20}, an agent believes an argument $\alpha$ to some degree if $P(\alpha) > 0.5$, disbelieves $\alpha$ to some degree if $P(\alpha)<0.5$, and neither believes nor disbelieves $\alpha$ when $P(\alpha) = 0.5$. 

\begin{definition}
    Let $\Pi$ be a value set, $\varphi= p(\alpha) \# v$ be an epistemic atom. The satisfying distributions of $\varphi$ are defined as $\mathrm{Sat}(\varphi) = \{P \in \mathrm{Dist}(G) \mid P(\alpha) \# v\}$, where $\mathrm{Dist}(G)$ is the set of all belief distributions on $\mathrm{Nodes}(G)$. The restricted satisfying distribution of $\varphi$ w.r.t. $\Pi$ is $\mathrm{Sat}(\varphi,\Pi) = \mathrm{Sat}(\varphi) \cap \mathrm{Dist}(G,\Pi)$, where $\mathrm{Dist}(G, \Pi)$ is the set of restricted distributions for a restricted value set $\Pi$.
    Let $\phi$ and $\psi$ be epistemic formulae. The set of restricted satisfying distributions for a given epistemic formula is as follows: $\mathrm{Sat}(\phi \wedge \psi) = \mathrm{Sat}(\phi) \cap \mathrm{Sat}(\psi)$; $\mathrm{Sat}(\phi \vee \psi)= \mathrm{Sat}(\phi) \cup \mathrm{Sat}(\psi)$; and $\mathrm{Sat}(\neg \phi)=\mathrm{Sat}(\top) \backslash \mathrm{Sat}(\phi)$ . For a set of epistemic formula $\Phi = \{\phi_{1},...,\phi_{n}\}$, the set of satisfying distribution is $\mathrm{Sat}(\Phi)=\mathrm{Sat}(\phi_{1}) \cap ... \cap \mathrm{Sat}(\phi_{n})$.
\end{definition}

\begin{example}
    Consider a restricted epistemic graph with nodes $\{A,B,C\}$, the epistemic atom $p(A)>0.5$ and the epistemic formula $p(A)>0.5 \rightarrow \neg(p(B)>0.5)$, where $\Pi=\{0,0.5,1\}$. An example of a belief distribution that satisfies this epistemic atom is $P_{1}$ s.t. $P_{1}(\emptyset)=0.2$, $P_{1}(\{A,B\})=0.3$ and $P_{1}(\{A\}) = 0.5$. The belief distribution $P_{2}$ s.t. $P_{2}(\emptyset)=1$ does not satisfy the epistemic atom. An example of a belief distribution that satisfies the epistemic formula is $P_{3}$ s.t. $P_{3}(\{A\})=0.2$, $P_{3}(\{A,B\})=0.4$ and $P_{3}(\{C\}) = 0.4$. The belief distribution $P_{4}$ s.t. $P_{4}(\{A\})=0.2$, $P_{4}(\{B\})=0.4$ and $P_{4}(\{A,B\}) = 0.4$ does not satisfy the epistemic formula.
\end{example}

The above epistemic constraints can be generated from a dataset, which is composed of a set of data items that are defined as follows.
\begin{definition}
    A data item is a function d from a set of arguments to a set of values. A dataset, $D$, is a set of data items over arguments. Data items reflect the beliefs of participants towards arguments.
\end{definition}

As mentioned above, the beliefs of agents are modeled in terms of probability distributions of an epistemic graph. Agents fill in the dataset according to their belief distributions that satisfy some epistemic atoms and epistemic formulae. For example, if the value of a data item of argument $\alpha$ equals $0.6$, then $P(\alpha)=0.6$, i.e., argument $\alpha$ is being believed by an agent. The corresponding epistemic atom is of the form $p(\alpha) = 0.6$.

Now, let us move on to the background knowledge for formulating a set of rationality principles. According to Dung's argumentation theory \cite{DBLP:journals/ai/Dung95}, given an abstract argumentation framework (AF for short), also called an argument graph, the AF can be mapped to a set of extensions according to an argumentation semantics that can be defined by a set of principles. The first principle is called the admissibility principle, which requires that each extension be conflict-free and each argument in the extension be defended by the extension. The second principle is called the reinstatement principle, which says that each argument accepted by an extension should belong to that extension. For more information about argumentation semantics and principles, the readers are referred to \cite{2011An}.

Under the context of probabilistic argumentation, Hunter and Thimm proposed a property called coherent property that satisfies the admissibility principle \cite{DBLP:journals/corr/HunterT14}, which is formally defined as follows. 

\begin{definition}
Let $AF =\langle A, \mathcal{R}_\mathrm{att}\rangle$ be an argument graph and $P:2^{A} \rightarrow [0,1]$, where $A$ is a set of arguments, and $\mathcal{R}_\mathrm{att} \subseteq A\times A$. P is coherent (COH) with respect to $AF$ if for all arguments $a,b \in A$ such that $a$ attacks $b$, then $P(a) \leq 1 - P(b)$.
\label{def_COH}
\end{definition}

For an epistemic graph that only exists attack relation, it can be viewed as an argument graph. In this case, in order for the epistemic constraints over the graph to be rational, it is desirable that the admissibility principle, the reinstatement principle and the coherent property can be satisfied.

For an epistemic graph that contains both attack and support relations, the rationality principles are related to deductive support and necessary support defined in a bipolar framework (called bipolar argument graph in this paper) \cite{DBLP:conf/ecsqaru/CayrolL05a,DBLP:journals/ijis/AmgoudCLL08,DBLP:conf/comma/BoellaGTV10,DBLP:conf/ictai/NouiouaR10,DBLP:conf/sum/NouiouaR11}.

\begin{definition}
Let $A$ be a finite and non-empty set of arguments, $\mathcal{R}_\mathrm{att}$ be a binary relation over $A$ called the attack relation and $\mathcal{R}_\mathrm{sup}$ be a binary relation over $A$ called the support relation.  A bipolar argument graph is represented as a 3-tuple $\langle A, \mathcal{R}_\mathrm{att}, \mathcal{R}_\mathrm{sup}\rangle$. 
\end{definition}

Deductive supports can be interpreted as follows \cite{DBLP:conf/comma/BoellaGTV10}:
Let $a$ and $b$ be two arguments. If $a\mathcal{R}_\mathrm{sup}b$, then the acceptance of $a$ implies the acceptance of $b$, and the non-acceptance of $b$ implies the non-acceptance of $a$. 

Necessary supports can be interpreted as follows \cite{DBLP:conf/ictai/NouiouaR10,DBLP:conf/sum/NouiouaR11}: Let $a$, $b$ be two arguments. If $a\mathcal{R}_\mathrm{sup}b$, then the acceptance of $a$ is necessary to obtain the acceptance of $b$, i.e., the acceptance of $b$ implies the acceptance of $a$. 

\section{A Rule Learning Framework Based On the 2-way Generalization Step}
\begin{figure}
\centering
\includegraphics[height=0.6in, width=2.6in]{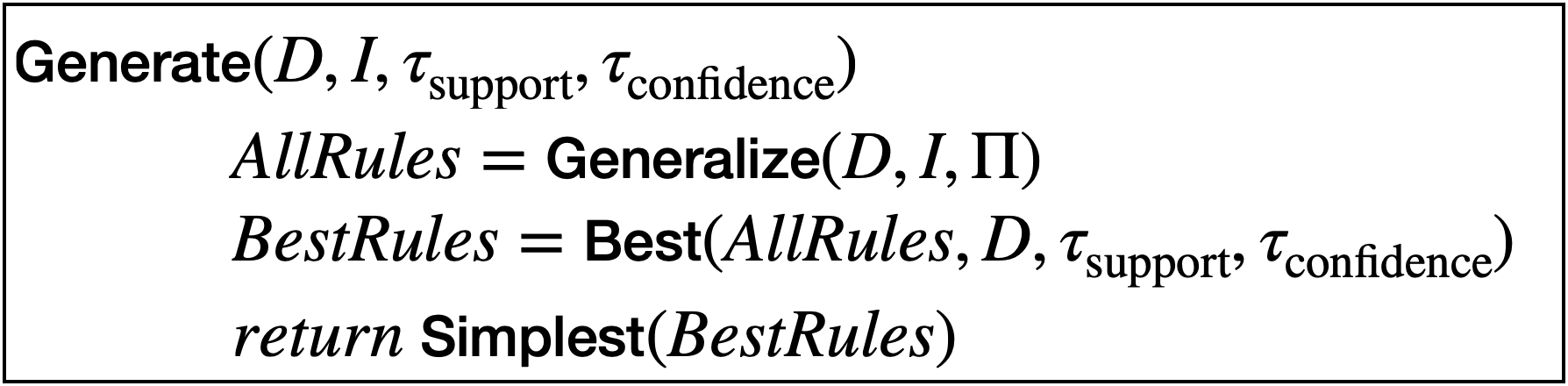}
\caption{The generating algorithm on dataset $D$.}
\label{fig_2way}
\end{figure}

Hunter \cite{DBLP:conf/comma/Hunter20} put forward a generating algorithm to generate a set of simplest and best rules from a dataset, as illustrated in Fig. \ref{fig_2way}.

This algorithm consists of the following three functions: $\mathrm{Generalize}(D,I,\Pi)$, $\mathrm{Best}(\textit{Rules},D,\tau_\mathrm{support},\tau_\mathrm{confidence})$ and $\mathrm{Simplest}(Rules)$. We first define the function $\mathrm{Generalize}(D,I,\Pi)$ as follows, based on notions of influence tuple and 2-way generalization step.
\begin{definition}
    Let $G$ be a graph, $\mathrm{Nodes}(G)$ be arguments in $G$. $\{\alpha_{1},...,\alpha_{n}\} \subseteq \mathrm{Nodes}(G)\backslash\{\beta\}$ and $\beta \in \mathrm{Nodes}(G)$. An influence tuple is a tuple $(\{\alpha_{1},...,\alpha_{n}\}, \beta)$, where each $\alpha_{i}$ influences $\beta$. We call each $\alpha_{i}$ an influencer and $\beta$ an influence target.
    \label{def_influenTuple}
\end{definition}

\begin{definition}
    Let $d$ be a data item, $I=(\{\alpha_1,...,\alpha_n\},\beta)$ be an influence tuple and $\Pi = \{0,0.5,1\}$ be a restricted value set. The following is a 2-way generalization step, where for each i, if $v_i>0.5$, then $\#_{i}$ is ``$>$", else if $v_i\leq 0.5$, then $\#_{i}$ is ``$\leq$".
    $$\frac {d(\alpha_{1})=v_{1},..., d(\alpha_{n})=v_{n}, d(\beta)=v_{n+1}}{p(\alpha_{1})\#_{1}0.5 \wedge ... \wedge p(\alpha_{n})\#_{n}0.5 \rightarrow p(\beta)\#_{n+1}0.5}$$
    If a data item $d$ satisfies the precondition (above the line), then a function $\mathrm{TwoWayGen}(d,I,\Pi)$ returns the rule given in the postcondition (below the line), otherwise it returns nothing.
\end{definition}

\begin{example}
    After applying the 2-way generalization step to the data in row $026$ of Table \ref{fig:ex1}, where the influence tuple is $(\{\textit{Dw6}, \textit{Dw2},\textit{Dw5}\}, \textit{Dw3})$, we obtain the following result: $p(\textit{Dw6}) \leq 0.5 \wedge p(\textit{Dw2})>0.5 \wedge p(\textit{Dw5}) \leq 0.5 \rightarrow p(\textit{Dw3})>0.5$. 
\end{example}

The function $\mathrm{Generalize}(D,I,\Pi)$ returns $\{\mathrm{TwoWayGen}(d,I,\Pi)|d\in D\}$, where $D$ is a dataset and $I$ is an influence tuple and $\Pi$ is a restricted value set.

Second, according to Hunter \cite{DBLP:conf/comma/Hunter20}, a generating algorithm can be constructed to generate rules that reflect beliefs in arguments among the majority of individuals in a dataset. If the belief distribution of a participant conflicts with the belief distributions of a certain percentage of individuals in a dataset, then the rule is regarded as counter-intuitive. Then, three parameters (which are \textit{Support}, \textit{Confidence} and \textit{Lift}) are set to judge whether a rule is intuitive, and if so, it is regarded as a best rule. 

 Let $\mathrm{Values}(p(\beta) \# v, \Pi)=\{x \in \Pi \mid x \# v\}$ be a set of values for an atom $p(\beta) \# v$ where $\Pi$ is a restricted value set. 

\begin{definition}
Let $d \in D$ be a data item, $G$ be a graph and $\Pi$ be a restricted value set. Let $R=\phi_{1} \wedge ... \wedge \phi_{n} \rightarrow \phi_{n+1}$ be a rule where $\phi_{i}$ is of the form $P(\alpha_{i}) \#_{i} v_{i}$ for $i \in \{1,...,n+1\}$, $v_{i} \in \Pi$ and $\alpha_{i} \in \mathrm{Nodes}(G)$. $R$ is \textbf{fired} by d iff for every $\phi_{i}$ s.t. $i \leq n$, $d(\alpha_{i}) \in \mathrm{Values}(p(\alpha_{i})\#_{i}v_{i},\Pi)$. $R$ \textbf{agrees} with $d$ iff $d(\alpha_{n+1}) \in \mathrm{Values}(\phi_{n+1},\Pi)$. $R$ is \textbf{correct} w.r.t. $d$ iff $R$ satisfies both fired and agrees conditions. Then, $\mathrm{Support}(R,D) = \frac{1}{|D|} \times | \{d \in D \mid R\ is\ fired\ by\ d\}|$, $\mathrm{Confidence}(R,D) = \frac{1}{|D|} \times |\{d \in D \mid R \ is\ correct\ w.r.t.\ d\}|$ and $\mathrm{Lift}(R,D) = \frac{|\{d \in D \mid R\ is\ correct\ w.r.t.\ d\}| \times |D|}{|\{d \in D \mid R\ is\ fired\ by\ d\}|\times|\{d \in D \mid R\ agrees\ with\ d\}|}$.
\label{def_para}
\end{definition}

\begin{example}
    Consider the rule $p(\textit{Dw2}) > 0.5 \rightarrow p(\textit{Dw6})<0.5$ with data from Table \ref{tab:ex1}. This rule is fired with row $026$, agrees with row $004$ and row $026$, and is correct with row $026$.
\end{example}

\begin{definition}
    Let $D$ be a data set, $\textit{Rules}$ be a set of rules, and $\tau_\mathrm{support} \in [0,1]$ (resp. $\tau_\mathrm{confidence} \in [0,1]$) be a threshold for support (resp. confidence). The set of \textbf{best rules} $\mathrm{Best}(\textit{Rules},D,\tau_\mathrm{support},\tau_\mathrm{confidence}) = \{R \in \textit{Rules} \mid \mathrm{Support}(R,D)>\tau_\mathrm{support}\ and\ \mathrm{Confidence}(R,D)>\tau_\mathrm{confidence}\ and\ \mathrm{Lift}(R,D)>1\}$. Let a rule $R = \phi_{1} \wedge ... \wedge \phi_{n} \rightarrow \psi$, $\mathrm{Conditions}(R) = \{\phi_{1},...,\phi_{n}\}$, $\mathrm{Head}(R)=\psi$, the set of \textbf{simplest rules} $\mathrm{Simplest}(\textit{Rules}) = \{R \in \textit{Rules} \mid for\ all\ R^{'} \in \textit{Rules},\ if\ \mathrm{Head}(R)=\mathrm{Head}(R^{'}),\ then\ \mathrm{Conditions}(R) \subseteq \mathrm{Conditions}(R^{'})\}$.
    \label{def_simplestbest}
\end{definition}

The generating algorithm proposed by Hunter generates rules based on the majority's beliefs in arguments. However, this is unsuitable when there exist public cognitive errors or most people are lying. Therefore, in practical applications, it is necessary to combine epistemic constraints with epistemic graphs when making decisions. Therefore, we intend to construct objective judging criteria to generate rational rules on the basis the of best rules.

\section{A Filtering-Based General Approach}
We now present a filtering-based general approach by introducing a multiple-way generalization step and rationality principles. This rule learning framework aims at obtaining a set of epistemic constraints such that the constraints are rational and their restricted values do not confine to 0.5, but also other values satisfying conditions of the restricted value set. We will first present standards to generate rational rules, and then give a formal definition of the learning problem as well as an algorithm for learning rules.

\subsection{Rationality principles and rational rules}
We define rational rules according to a set of rationality principles for three situations of a bipolar argument graph: when a graph only contains attack relation, when a graph only contains support relation and  when a graph contains both attack and support relations. 

The rationality principles for defining rational rules are based on Dung's argumentation theory (for attack relation) and the deductive and necessary supports (for support relations) introduced in Section 2. 

\begin{definition}
Let  $\langle A, \mathcal{R}_\mathrm{att}, \mathcal{R}_\mathrm{sup}\rangle$ be a bipolar argument graph where $\mathcal{R}_\mathrm{sup}$ is empty,  and $P:2^\mathrm{A} \rightarrow [0,1]$ be a probability distribution over $A$. Let $a \in A$ be an influence target, and $\mathrm{Att}(a) \subseteq A$ be a set of influencers that attack $a$. The following two rationality principles w.r.t. $P$ are defined to identify irrational rules:
\begin{itemize}
    \item[$C_{1}$] $P$ is incoherent if there exists an argument $b$ in $\mathrm{Att}(a)$ such that b is believed and $a$ is believed simultaneously.
    \item[$C_{2}$] $P$ is non-reinstated if all arguments in $\mathrm{Att}(a)$ are disbelieved and $a$ is also disbelieved.
\end{itemize}
\label{attack_rational}
\end{definition}

In this definition, Principle $C_{1}$ is derived from the coherent property mentioned in Definition \ref{def_COH}, and Principle $C_{2}$ is derived from the reinstatement principle. A rule satisfying any one of the principles is regarded as irrational.

\begin{example}
Consider Table \ref{table_attack}. $\textit{Sys4}$ denotes ``Italy is the best place to live", $\textit{Sys7}$ denotes ``Our society gets worse and worse by year" and $\textit{Dw6}$ denotes ``The society has become more lawless and bestial nowadays". According to the text descriptions, we regard $\textit{Sys7}$ and $\textit{Dw6}$ as attackers of $\textit{Sys4}$, and the corresponding influence tuple is $I=(\{\textit{Sys7}, \textit{Dw6}\},\textit{Sys4})$. Let the restricted value set $\Pi = \{0,0.5,1\}$. Consider row $001$, we can obtain a rule $p(\textit{Sys7})<0.5 \wedge p(\textit{Dw6})>0.5 \rightarrow p(\textit{Sys4}) > 0.5$. The attacker $\textit{Dw6}$ is believed (i.e. $d_{001}(\textit{Dw6})>0.5$) and the influence target $\textit{Sys4}$ is believed (i.e. $d_{001}(\textit{Sys4})>0.5$) simultaneously. This rule satisfies Principle $C_{1}$, and is regarded as irrational. Now consider row $000$, we can obtain a rule $p(\textit{Sys7})<0.5 \wedge p(\textit{Dw6})<0.5 \rightarrow p(\textit{Sys4})<0.5$. Both attackers $\textit{Sys7}$ and $\textit{Dw6}$ are disbelieved, and the influence target $\textit{Sys4}$ is also disbelieved. This rule satisfies Principle $C_{2}$, and is regarded as irrational.
\end{example}

\begin{definition}
Let  $\langle A, \mathcal{R}_\mathrm{att}, \mathcal{R}_\mathrm{sup}\rangle$ be a bipolar argument graph where $\mathcal{R}_\mathrm{att}$ is empty,  and $P:2^\mathrm{A} \rightarrow [0,1]$ be a probability distribution over $A$. Let $a \in A$ be an influence target, and $\mathrm{Sup}(a) \subseteq A$ be a set of influencers that support $a$. The following two rationality principles w.r.t. $P$ are defined to identify irrational rules:
\begin{itemize}
    \item[$C_{3}$] $P$ is non-conclusive if there exists an argument $b$ in $\mathrm{Sup}(a)$ such that $b$ is believed but $a$ is disbelieved.
    \item[$C_{4}$] $P$ is non-grounded if all arguments in $\mathrm{Sup}(a)$ are disbelieved but $a$ is believed.
\end{itemize}
 \label{support_rational}
\end{definition}

Principle $C_3$ is based on the nature of deductive support, while Principle $C_4$ is based on the nature of necessary support. In this paper, we assume that any rule satisfying one of these two principles is irrational. 

\begin{table*}
\begin{floatrow}
\capbtabbox{
\resizebox{40mm}{8mm}{
\begin{tabular}{|c|c|c|c|c|}
\hline
& Sys4 & Sys7 & Dw6 \\
\hline
000 & 0.2 & 0.3 & 0.3 \\
001 & 0.6 & 0.3 & 0.6 \\
\hline
\end{tabular}
}
}{
 \caption{Containing columns and rows of data obtained from the Italian study \cite{PELLEGRINI2019104144}.}
 \label{table_attack}
}
\capbtabbox{
\resizebox{40mm}{8mm}{
\begin{tabular}{|c|c|c|c|c|}
\hline
& Qu1 & Im1 & Im2 \\
\hline
001 & 0.7 & 0.1 & 0.2 \\
002 & 0.3 & 0.3 & 0.7 \\
\hline
\end{tabular}
}
}{
 \caption{Containing columns and rows of data obtained from the Spanish study \cite{asi.23488}.}
 \label{table_support}
}
\end{floatrow}
\end{table*}

\begin{example}
Consider Table \ref{table_support}. $\textit{Im1}$ denotes ``The use of Wikipedia is well considered among colleagues", $\textit{Im2}$ denotes ``My colleagues use Wikipedia" and $\textit{Qu1}$ denotes ``Articles in Wikipedia are reliable". According to the text descriptions, we regard $\textit{Im1}$ and $\textit{Im2}$ as supporters of $\textit{Qu1}$, and the corresponding influence tuple is $I=(\{\textit{Im1}, \textit{Im2}\},\textit{Qu1})$. Let the restricted value set $\Pi=\{0,0.5,1\}$. The rule $p(\textit{Im1})<0.5 \wedge p(\textit{Im2})>0.5 \rightarrow p(\textit{Qu1})<0.5$ obtained from row $002$ is irrational since it satisfies Principle $C_{3}$. The rule $p(\textit{Im1})<0.5 \wedge p(\textit{Im2})<0.5 \rightarrow p(\textit{Qu1})>0.5$ obtained from row $001$ is also irrational since it satisfies Principle $C_{4}$.
\end{example}

\begin{definition} 
Let  $\langle A, \mathcal{R}_\mathrm{att}, \mathcal{R}_\mathrm{sup}\rangle$ be a bipolar argument graph,  and $P:2^\mathrm{A} \rightarrow [0,1]$ be a probability distribution over $A$. Let $a \in A$ be an influence target, and $\mathrm{Att}(a)$ and $\mathrm{Sup}(a)$ be two sets of influencers that attack and support $a$ respectively, where $\mathrm{Att}(a)\cap \mathrm{Sup}(a) = \emptyset$. The following two rationality principles w.r.t. $P$ are defined to identify irrational rules:
\begin{itemize}
    \item[$C_{5}$] $P$ is gen-nonconclusive if all attackers in $\mathrm{Att}(a)$ are disbelieved but there exists a supporter $b$ in $\mathrm{Sup}(a)$ such that $b$ is believed, and $a$ is disbelieved at the same time.
    \item[$C_{6}$] $P$ is gen-nongrounded if all supporters in $\mathrm{Sup}(a)$ are disbelieved and there exists an attacker b in $\mathrm{Att}(a)$ such that $b$ is believed, and $a$ is believed at the same time.
\end{itemize}
\label{bi_rational}
\end{definition}

Principles $C_{5}$ and $C_{6}$ are a generalization of Principles $C_{3}$ and $C_{4}$ respectively.

\begin{example}
    Consider Table \ref{tab:ex1}. According to the text descriptions, we regard $\textit{Dw2}$ and $\textit{Dw3}$ as supporters of $\textit{Dw6}$, and $\textit{Dw5}$ as an attacker of $\textit{Dw6}$. The corresponding influence tuple is $I=(\{\textit{Dw2}, \textit{Dw3},\textit{Dw5}\},\textit{Dw6})$. Let the value set $\Pi = \{0,0.5,1\}$. The rule $p(\textit{Dw2}) > 0.5 \wedge p(\textit{Dw3}) > 0.5 \wedge p(\textit{Dw5}) < 0.5 \rightarrow p(\textit{Dw6}) < 0.5$ obtained from row $026$ is irrational since it satisfies Principle $C_{5}$. The rule $p(\textit{Dw2}) < 0.5 \wedge p(\textit{Dw3}) < 0.5 \wedge p(\textit{Dw5}) > 0.5 \rightarrow p(\textit{Dw6}) > 0.5$ obtained from row $111$ is irrational since it satisfies Principle $C_{6}$.
\end{example}

It is essential to represent the nature of relations when filtering rational rules. We use a relation item to represent the attack and support relations between an influencer and an influence target.
\begin{definition}
    Let $I=(\{\alpha_{1},...,\alpha_{n}\},\beta)$ be an influence tuple, and A be a set of arguments. A relation item is a partial function $\mathrm{r}:A \times A \rightarrow \{0,1\}$, where ``$0$" represents the attack relation and ``$1$" represents the support relation. So for $\alpha \in \{\alpha_{1},...,\alpha_{n}\}$, $\mathrm{r}(\alpha,\beta) \in \{0,1\}$. A relation set $\mathrm{Rel}(I)=(\textit{r}_{1},...,\textit{r}_{n})$ is a set containing all the relation items between the influencers and the influence target of $\textit{I}$.
\end{definition}

\begin{example}
    The following shows some arguments from the politics database \cite{PELLEGRINI2019104144}.\\
    (\textit{Dw2}) Chaos and anarchy can erupt around us nowadays.\\
    (\textit{Dw3}) There exist dangerous people in current society.\\
    (\textit{Dw4}) There isn't any more crime in the street" nowadays.\\
    (\textit{Dw5}) The society isn't dangerous nowadays.\\
    Based on these text descriptions, we regard $\textit{Dw2}$ and $\textit{Dw3}$ as attackers of $\textit{Dw5}$, and $\textit{Dw4}$ as a supporter of $\textit{Dw5}$. $\textit{I} = (\{\textit{Dw2}, \textit{Dw3}, \textit{Dw4}\},\textit{Dw5})$ is the corresponding influence tuple. Therefore, $\mathrm{r}(\textit{Dw2},\textit{Dw5})=0$, $\mathrm{r}(\textit{Dw3},\textit{Dw5})=0$ and $\mathrm{r}(\textit{Dw4},\textit{Dw5})=1$. The relation set $\mathrm{Rel}(I)=(0,0,1)$.
\end{example}

Given a set of rules \textit{Rules} and an influence tuple $I$ and a set of relation items $\mathrm{Rel}(I)$, rational rules are those that do not meet any rationality principle. We denote the set of rational rules as $\mathrm{Rational}(\textit{Rules},I,\mathrm{Rel}(I))$.


\subsection{A multiple-way generalization step}
Now we give a formal definition of the learning problem:\\
\textbf{Given}
\begin{itemize}
    \item [1.] A dataset $D=\{d_{1},...,d_{n}\}$ comes from a study that uses Likert scales to record responses of participants towards arguments, where $d_{i} \in D$ is a data item. We use the 11-point scale to map each value of the dataset to a probability value. As a result, for a set of arguments $A$ and an argument $\alpha \in A$, $d(\alpha) \in \{0, 0.1, 0.2,...,0.9,1\}$. For more information about the Likert scale and the 11-point scale, readers are referred to Likert \cite{1932A} and Hunter \cite{DBLP:conf/comma/Hunter20}. 
    \item[2.] An influence tuple $I$, which is constructed by hand according to the text descriptions of arguments. Note that we need at least two arguments to form an influence tuple. Based on $I$, a set of relation items $\mathrm{Rel}(I)$ can be developed where $\mathrm{Rel}(I) \neq \emptyset$.
    \item[3.] A restricted value set $\Pi$.
    \item[4.] Thresholds $\tau_{\mathrm{confidence}} \in [0,1]$ and $\tau_{\mathrm{support}}\in [0,1]$.
\end{itemize}

\noindent \textbf{Find} a set of rules $Rules$ that is selected from the following set of candidate rules where $I=(\{\beta_{1},...,\beta_{n}\},\alpha)$:
\begin{eqnarray*}
\mathrm{Rules}(I,\Pi)&=&\{p(\gamma_{1}) \#_{1} v_{1} \wedge ... \wedge p(\gamma_{k}) \#_{k} v_{k} \rightarrow p(\alpha) \#_{k+1} v_{k+1}\ \mid \\
&& \{\gamma_{1},...,\gamma_{k}\} \subseteq \{\beta_{1},...,\beta_{n}\},  \#_{i} \in \{<,>,\leq,\geq\},\ and\  v_{i} \in \Pi \backslash \{0,1\}\}
\end{eqnarray*}
Rules generated from a dataset might be inconsistent with their epistemic graphs. These rules are considered as irrational. We want to generate a set of simplest and best rational rules $R$ s.t.:
\begin{itemize}
    \item[1.] $\forall r \in \textit{R}$, $r \in \textit{Rational}$ where $\textit{Rational} = \mathrm{Rational}(\textit{Rules}, I,\mathrm{Rel}(I))$.
    \item[2.] $\forall r \in R$, $r \in \textit{Best}$ where $\textit{Best} = \mathrm{Best}(\textit{Rational},D,\tau_\mathrm{support},\tau_\mathrm{confidence})$.
    \item[3.] $\forall r \in R$, $r \in \mathrm{Simplest}\textit{(Best)}$.
\end{itemize}

We use a multiple-way generalization step to generate rules from a dataset. Different from the 2-way generalization step, this method is capable of representing beliefs by using tighter intervals, and therefore can better reflect participants' beliefs. For example, for the data item $d(\alpha)=0.2$ where $\alpha$ is an argument, it is better to represent the epistemic atom as $p(\alpha) \leq 0.25$ (which stands for strongly disbelieves $\alpha$) instead of $p(\alpha)<0.5$ (which stands for disbelieves $\alpha$). In order to better reflect agents' beliefs, it is essential for us to avoid generating epistemic atoms such as $p(\alpha)>0.25$, from which we cannot decide whether an agent believes argument $\alpha$ or not. To address this problem, we introduce a new notion called the \textsf{Nearest} function defined as follows.

\begin{definition}
    Let $\Pi$ be a restricted value set, $d$ be a data item, $\alpha$ be an argument, $l(d(\alpha),\pi)$ be the distance between $d(\alpha)$ and $\pi$, where $\pi \in \Pi$. The \textsf{Nearest} function $\mathrm{N}(d(\alpha),\Pi)$ is defined as follows: 
    \begin{itemize}
        \item[]  If $d(\alpha) \leq 0.5$, $\mathrm{N}(d(\alpha),\Pi) = \mathop{argmin}\limits_{\pi \in \Pi\ s.t.\ d(\alpha) \leq \pi \leq 0.5}\{l(d(\alpha),\pi)\}$; 
        \item[]  If $d(\alpha)> 0.5$, $\mathrm{N}(d(\alpha),\Pi)=\mathop{argmin}\limits_{\pi \in \Pi\ s.t.\ 0.5 \leq \pi \leq d(\alpha)}\{l(d(\alpha),\pi)\}$.
    \end{itemize}
\end{definition}

\begin{example}
    Let $\Pi = \{0,0.25,0.5,0.75,1\}$ and $\alpha$ be an argument. If $d(\alpha) = 0.3$, then $\mathrm{N}(0.3,\Pi) = 0.5$. If $d(\alpha) = 0.8$, then $\mathrm{N}(0.8,\Pi)= 0.75$. 
\end{example}

\begin{definition}
The multiple-way generalization step is divided into two sub steps, and is defined as follows:
Let $d$ be a data item and $I = (\{\alpha_{1},...,\alpha_{n}\},\beta)$ be an influence tuple. The first step is the following.
\begin{eqnarray*}
\frac{d(\alpha_{1})=v_{1},...,d(\alpha_{n})=v_{n},d(\beta)=v_{n+1}}{p(\alpha_{1})= v_{1} \wedge...\wedge p(\alpha_{n})=v_{n} \rightarrow p(\beta)= v_{n+1}}
\end{eqnarray*}
If a data item $d$ satisfies the precondition (above the line), then $\mathrm{PreGen}(d,I)$ returns the rule given in the postcondition (below this line), otherwise it returns nothing.

Let $\Pi$ be a restricted value set. For each $i$, if $v_{i} > \mathrm{N}(v_{i},\Pi)$, then $\#_{i}$ is ``$>$"; if $v_{i} < \mathrm{N}(v_{i},\Pi)$, then $\#_{i}$ is ``$<$"; if $v_{i} = \mathrm{N}(v_{i},\Pi)$ and $v_{i} > 0.5$, then $\#_{i}$ is ``$\geq$"; If $v_{i} = \mathrm{N}(v_{i},\Pi)$ and $v_{i} < 0.5$, then $\#_{i}$ is ``$\leq$". The second step is defined as follows.


\begin{eqnarray*}
\frac{p(\alpha_{1})= v_{1} \wedge...\wedge p(\alpha_{n})=v_{n} \rightarrow p(\beta)= v_{n+1}}{p(\alpha_{1})\#_{1}\mathrm{N}(v_{1},\Pi) \wedge...\wedge p(\alpha_{n})\#_{n}\mathrm{N}(v_{n},\Pi) \rightarrow p(\beta)\#_{n+1}\mathrm{N}(v_{n+1},\Pi)} 
\end{eqnarray*}
If the rule $r$ in the precondition is satisfied, then $\mathrm{MultiWayGen}(r,\Pi)$ returns the rule given in the postcondition, otherwise it returns nothing.
\end{definition}

\begin{example}
    Let $\Pi = \{0,0.25,0.5,0.75,1\}$ be a restricted value set and $I = (\{\textit{Dw2},\textit{Dw5},\textit{Dw3}\},\textit{Dw6})$ be an influence tuple. By applying the first step of the multiple-way generalization step to row $004$ in Table \ref{tab:ex1}, we obtain the rule $p(\textit{Dw2}) = 0.3 \wedge p(\textit{Dw5}) = 0.3 \wedge p(\textit{Dw3}) = 0.3 \rightarrow p(\textit{Dw6}) = 0.2$. This rule is regarded as rational since it doesn't satisfy any rationality principle. Now, we can move on to applying the second step to this rule. 
    \begin{eqnarray*}
    p(\textit{Dw2}) < 0.5 \wedge p(\textit{Dw5}) < 0.5 \wedge p(\textit{Dw3}) < 0.5 \rightarrow p(\textit{Dw6}) < 0.25\\\
    \end{eqnarray*}
\end{example}

\begin{definition}
    Let $D$ be a dataset, $I$ be an influence tuple, and $\Pi$ be a restricted value set. The generalize function $\mathrm{PreGeneralize}(D,I)$ returns the set $\{\mathrm{PreGen}(d,I) \mid d \in D\}$, and the generalize function $\mathrm{GeneralizeMulti}(R,\Pi)$ returns the set $\{\mathrm{MultiWayGen}(r,\Pi)\mid r \in R\}$, where $R$ is a set of rules generated from $\mathrm{PreGeneralize}(D,I)$ after filtering out a set of irrational rules.
\end{definition}

\begin{figure}
\centering
\includegraphics[height=1in, width=3.2in]{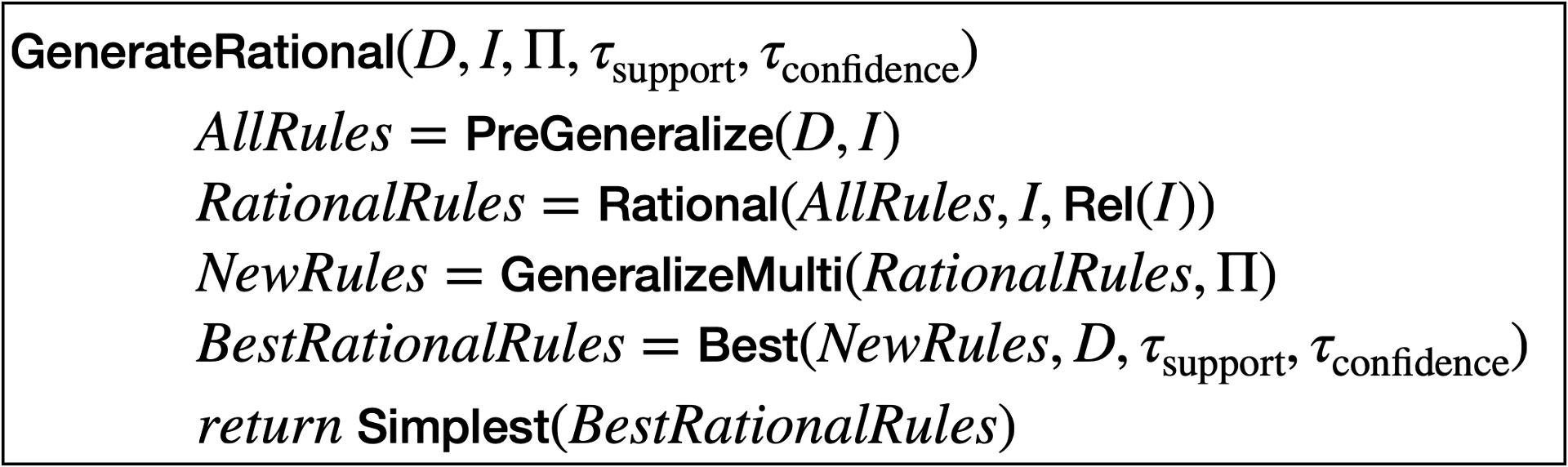}
\caption{The generating algorithm based on the multiple-way generalization step to generate rational rules.} \label{fig_Algorithm_Im}
\end{figure}

The improved algorithm for generating rational rules is given in Fig. \ref{fig_Algorithm_Im}.

\section{Empirical Study}
In this paper, we consider crowd-sourcing data from two studies, i.e., views on political attitudes in Italy \cite{PELLEGRINI2019104144} and the appropriateness of Wikipedia in teaching in Higher Education \cite{asi.23488}, that use the Likert scale to record users' beliefs in arguments. In the two studies, each statement is regarded as an argument, and each row in the data reflects the probability distribution of a user's beliefs (as shown in Table \ref{tab:ex1}).

To evaluate our approach on two datasets, we split each dataset into a training dataset and a testing dataset by randomly selecting 80 percent and 20 percent of the dataset respectively. Moreover, we set a maximum of four conditions per rule to avoid over-fitting. For inputs of the approach, we construct a set of influence tuples and a relation set by hand according to the statements in the studies, and choose a restricted value set that meets our needs. Following are examples of some rules generated from the Spanish and Italy studies.

\begin{example}
    Let $\Pi = \{0,0.25,0.5,0.75,1\}$, $\mathrm{Rel}(I)=(0,0,0,0,0,1,0)$ and $I= (\{\textit{Sys1},\textit{Sys2},\textit{Sys4},\textit{Sys5},\textit{Sys6},\textit{Sys7},\textit{Sys8}\},\textit{Sys3})$. The following exemplifies the rules learnt from the dataset of the Italian study.
    \begin{itemize}
        \item[1.] $p(\textit{Sys5}) \leq 0.25 \wedge p(\textit{Sys2}) \leq 0.25 \wedge p(\textit{Sys1}) \leq 0.5 \rightarrow p(\textit{Sys3}) > 0.5$ 
        \item[2.] $p(\textit{Sys7}) > 0.5 \wedge p(\textit{Sys1}) \leq 0.5 \rightarrow p(\textit{Sys3})>0.5$
        \item[3.] $p(\textit{Sys7}) > 0.5 \wedge p(\textit{Sys2}) \leq 0.25 \rightarrow p(\textit{Sys3}) > 0.5$
    \end{itemize}
\end{example}

\begin{example}
    Let $\Pi = \{0,0.25,0.5,0.75,1\}$, $I= (\{\textit{JR1},\textit{JR2},\textit{SA1},\textit{SA2},\textit{SA3},\textit{Im1},$  $\textit{Im2},\textit{Pf1},\textit{Pf2},\textit{Pf3},\textit{Qu1},\textit{Qu2},\textit{ENJ1}\},\textit{Qu3})$ and $\mathrm{Rel}(I)=(1,1,1,1,1,1,1,1,1,1,\\1,1,1)$. The following are some rules learnt from the dataset of the Spanish study.
    \begin{itemize}
        \item[1.] $p(\textit{Pf3}) \leq 0.5 \wedge p(\textit{Im1}) \leq 0.5 \rightarrow p(\textit{Qu3}) \leq 0.5$ 
        \item[2.] $p(\textit{Pf3}) \leq 0.5 \wedge p(\textit{Im2}) \leq 0.5 \rightarrow p(\textit{Qu3})\leq 0.5$
        \item[3.] $p(\textit{Qu1}) \leq 0.5 \rightarrow p(\textit{Qu3}) \leq 0.5$
        \item[4.] $p(\textit{Qu2}) \leq 0.5 \rightarrow p(\textit{Qu3}) \leq 0.5$
    \end{itemize}
\end{example}

Our algorithm is being implemented using the VS Code of Python 3.8.5 version and evaluated on Ubuntu gcc 7.5.0 with Intel(R) Xeon(R) Gold 6330 2GHz processor and 256GB RAM.\footnote{Code available at: https://github.com/cx3506/CLAR.git.} Results for evaluation are obtained from implementing the algorithm with ten repetitions, and we set $\tau_\mathrm{support}$ as $0.4$ and $\tau_\mathrm{confidence}$ as $0.8$. 

\begin{figure}
\centering
\includegraphics[height=2.4in, width=4.65in]{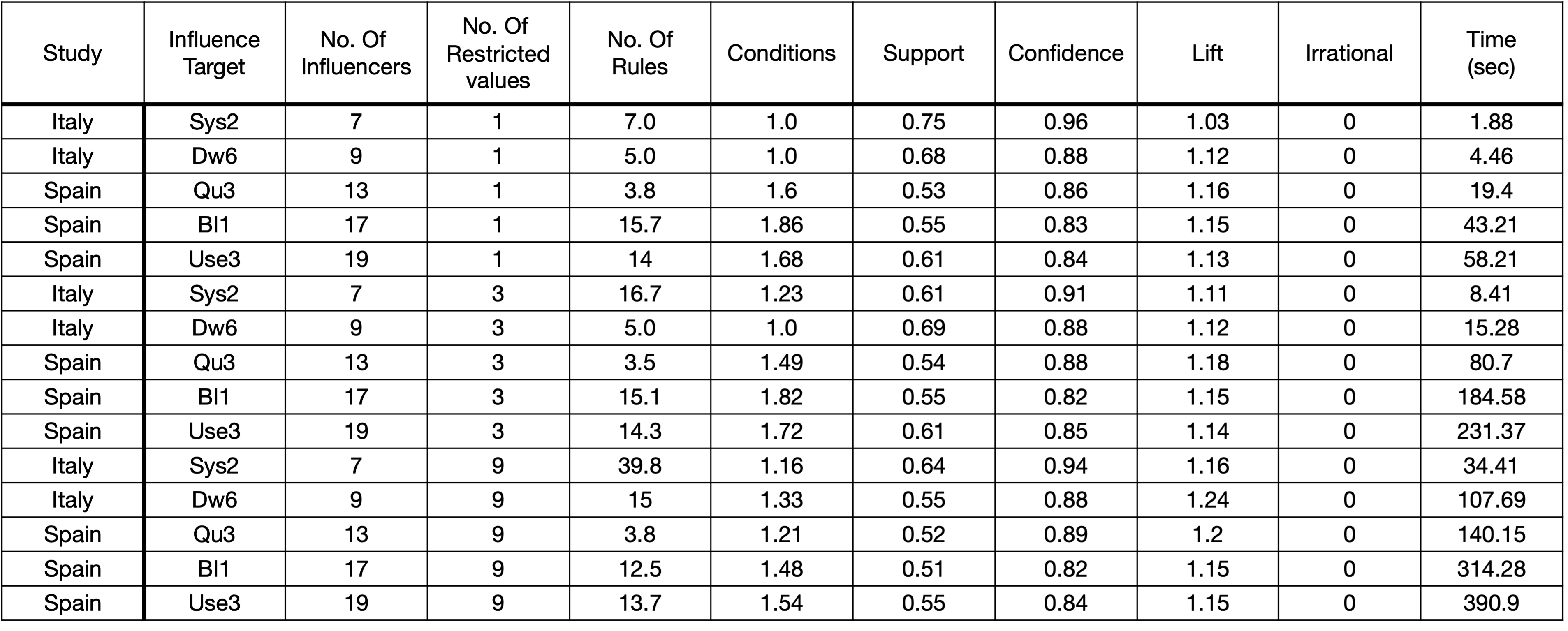}
\caption{Results for the Spanish and Italian datasets after implementing the filtering-based approach by setting restricted value set as $\{0,0.5,1\}$, $\{0,0.25,0.5,0.75,1\}$ and $\{0,0.1,0.2,0.3,0.4,0.5,0.6,0.7,0.8,0.9,1\}$ respectively. Column 4 presents the number of values in a restricted value set except for ``$0$" and ``$1$". Column 5 presents the number of generated rules. Column 6 presents the average number of conditions per rule. Column 10 presents the number of irrational rules among the learnt rules. For an approach to be qualified for filtering rational rules, the results of Column 10 should equal $0$.} \label{Table_improve}
\end{figure}

\begin{figure}[h!]
\centering
\includegraphics[height=1.3in, width=4.5in]{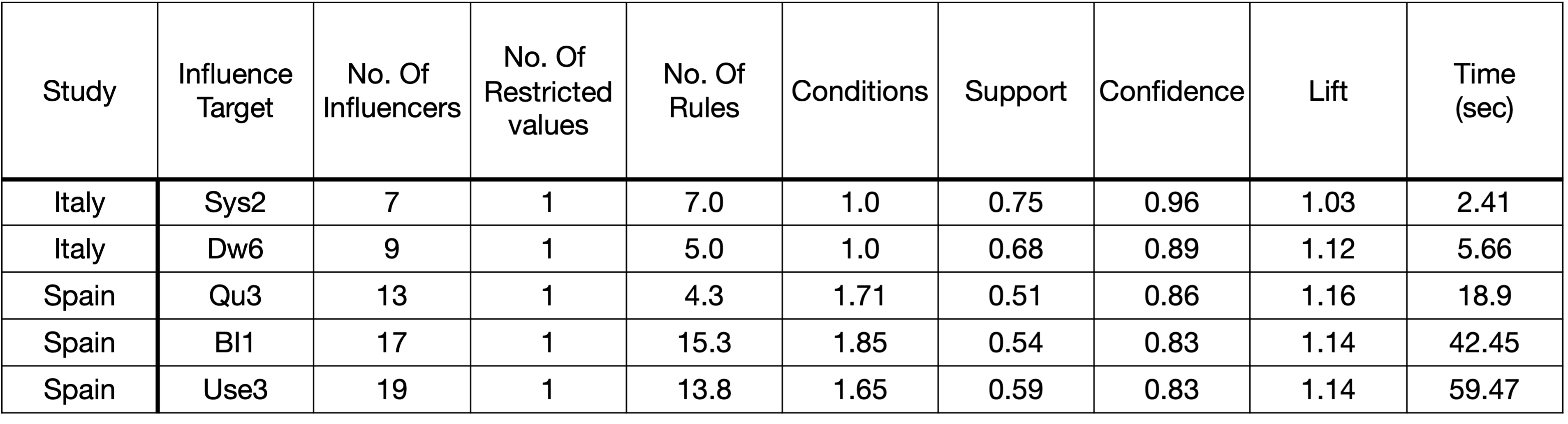}
\caption{Results for the Spanish and Italian datasets after implementing the framework proposed by Hunter \cite{DBLP:conf/comma/Hunter20} by setting restricted value set as $\{0,0.5,1\}$.} \label{Table_Ori}
\end{figure}

Results in the table of Fig. \ref{Table_improve} show that the filtering-based approach using a multiple-way generalization step performs well when generating a wider variety of rules since quality constraints from data (i.e., \textit{Confidence}, \textit{Support}, \textit{Lift} and \textit{Irrational}) are reasonable, where $\textit{Support} > 0.4$, $\textit{Confidence} > 0.8$, $\textit{Lift} > 1$ and $\textit{Irrational} = 0$. The number of rules generated for each influence target is quite large, indicating that a target is influenced by its influencers in a number of ways, which provides the APS with plenty of information to persuade a user to believe the target. In addition, by comparing the table in Fig. \ref{Table_improve} with the table in Fig. \ref{Table_Ori}, we discover that our new approach has similar performance in terms of $\textit{Lift}$, $\textit{Support}$ and $\textit{Confidence}$. This demonstrates from another perspective that our algorithm performs well. Meanwhile, the number of generated rules of the two algorithms are similar. This implies that most users' beliefs coincide with the rationality encoded in epistemic graphs.

Now we move on to see the time performance of our approach by comparing it with the framework proposed by Hunter \cite{DBLP:conf/comma/Hunter20}. Following are two graphs presenting the consuming time of learning rules using the original algorithm and our new algorithm.

\begin{figure}[htbp]
    \begin{minipage}[t]{0.5\linewidth}
	\centering
    \includegraphics[height=1.8in, width=2.4in]{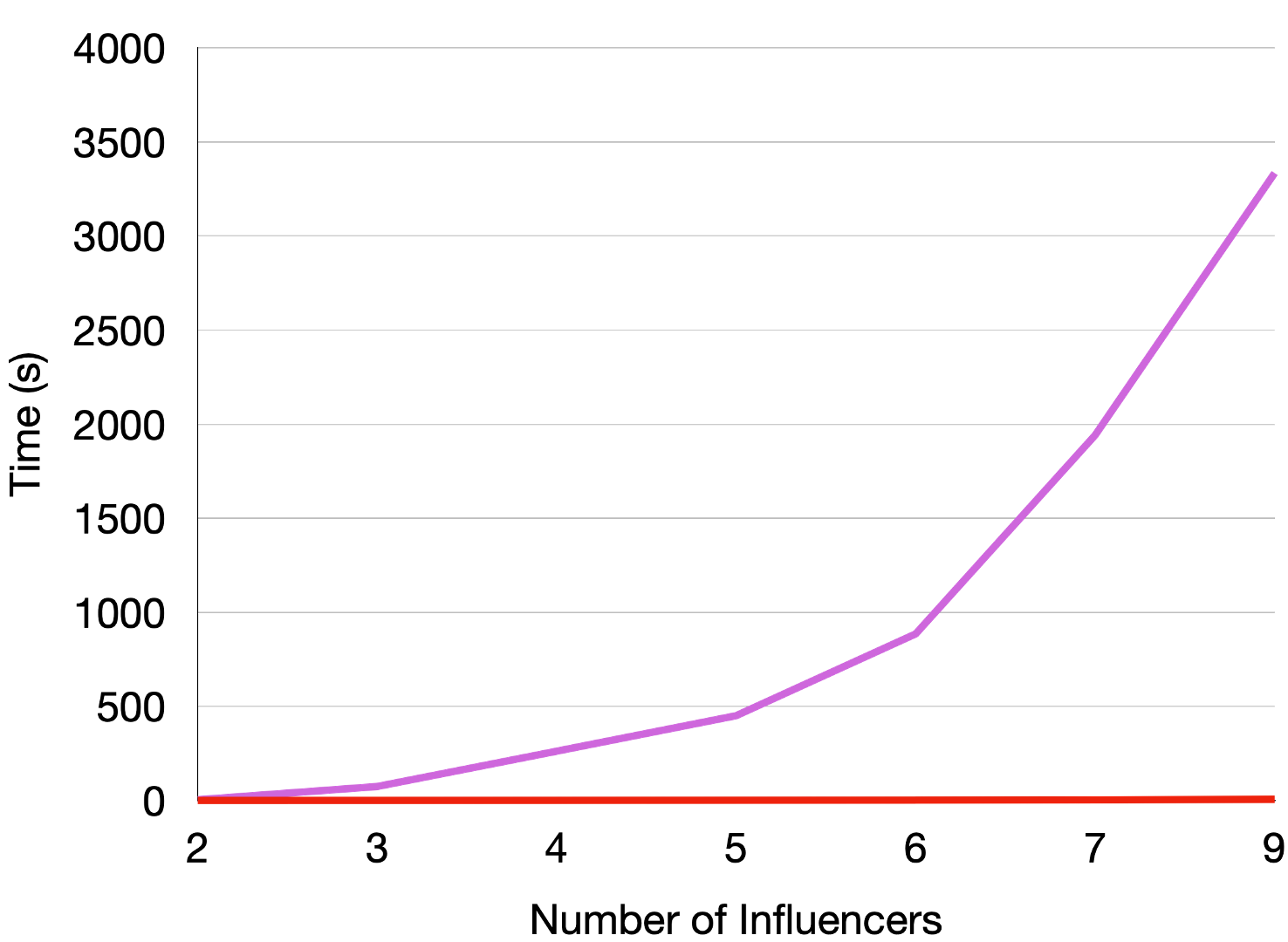}
    \end{minipage}%
    \begin{minipage}[t]{0.5\linewidth}  
    \centering
    \includegraphics[height=1.8in, width=2.4in]{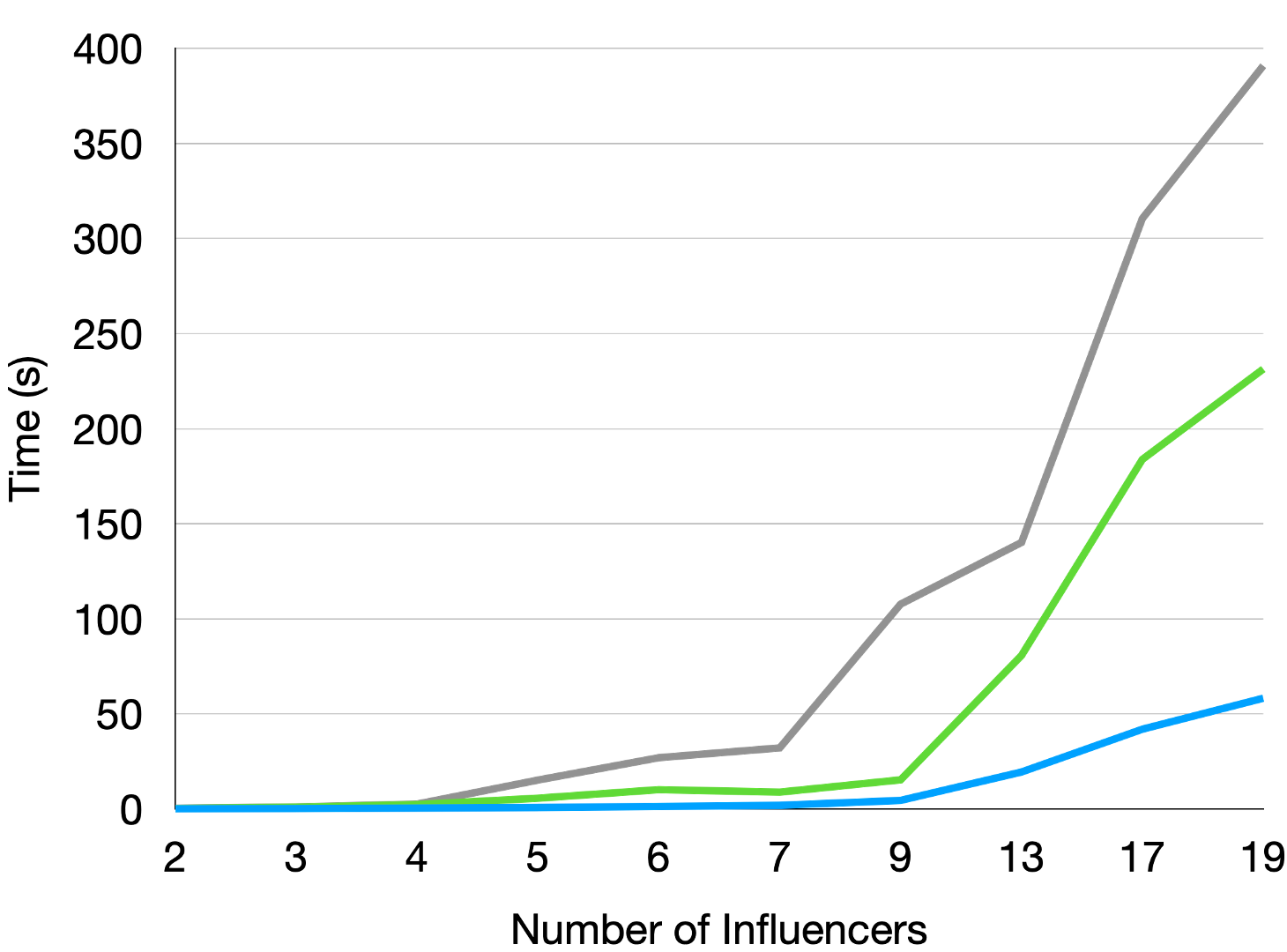} 
    \end{minipage}
    \caption{The left line chart presents the consuming time of the original framework \cite{DBLP:conf/comma/Hunter20} when generating rules on two different restricted value sets. The right line chart presents the consuming time of the filtering-based approach when generating rules on three different restricted value sets.}
    \label{Pic_time}
\end{figure}

On the left part of Fig. \ref{Pic_time}, the purple line and the red line represent the situation when $\Pi = \{0,0.25,0.5,0.75,1\}$ and $\Pi = \{0,0.5,1\}$ respectively. It indicates that the consuming time increases exponentially as the number of values of the restricted value set increases, leading to an unacceptable performance when generating a wider variety of rules. Now we consider applying the filtering-based approach to generating rules from the Spanish and Italian datasets. On the right part of Fig. \ref{Pic_time}, the gray line, the green line and the blue line represent the situations when $\Pi = \{0,0.1,0.2,0.3,0.4,0.5,0.6,0.7,0.8,0.9,1\}$, $\Pi = \{0,0.25,0.5,0.75,1\}$ and $\Pi = \{0,0.5,1\}$ respectively. Combining this graph with the table in Fig. \ref{Table_improve}, we observe that the time performance is acceptable (less than 400s) when dealing with a quite large number of influencers (up to 19) with multiple restricted values. In other words, our approach made significant progress when generating a wider variety of rules compared to the original framework proposed by Hunter \cite{DBLP:conf/comma/Hunter20}.

\section{Conclusions and Related Work}
In this paper, we have formulated an efficient filtering-based approach to generate a wider variety of rules from a dataset using a multiple-way generalization step. The main contributions of this paper are three-fold. First, we put forward six rationality principles, based on which the resulting algorithm can generate the best and rational rules that reflect information in both the domain model and the user model, and therefore can be directly harnessed by the APS strategy. Second, we developed a  multiple-way generalization step with the notion of a new generalize function such that various degrees of users' beliefs can be represented. Third, by proposing a \textsf{Nearest} function, we made remarkable progress in improving the efficiency of computation when expanding the variety of rules while maintaining a good performance in quality constraints.



Our algorithm is related to the work on rule learning. Inductive logic programming (ILP) by Muggleton \cite{muggleton1991inductive} is a well-known symbolic rule learning method that induces a set of logical rules by generalizing background knowledge and a given set of training examples, i.e., positive examples and negative examples. Our approach and ILP share a common need for background knowledge, but the types of the knowledge are different. The former uses epistemic graphs, while the latter uses first-order formulae.  

Hunter et al. \cite{DBLP:journals/ai/HunterPT20} set forth a key application of epistemic graphs, which is the automated persuasion system (APS). The APS persuades a user to believe something by offering arguments that have a high probability of influencing the user. When selecting moves in persuasion dialogues, a choice needs to be made between relying on either the domain model (the epistemic graph) or the user model (epistemic constraints generated from crowd-sourcing data). Our framework is able to generate rules which can reflect both information in the domain model and the user model. 

However, in this paper, we haven't considered how we can apply the learnt rules to an automated persuasion system. Moreover, it remains a question of how to generate more specific rational rules for different fields, where background knowledge varies. In the future, we want to put our filtering-based approach into practice, such as persuading users who have different areas of expertise. For application in a specific field, we may set forward principles based on background knowledge of the field. These principles might supplement the rationality principles proposed in this paper. Besides applying this approach into practice, there exists an unresolved problem with the existing approach, which is deciding influence tuples and relations between arguments by hand. To deal with this problem, one may refer to argument mining which can automatically create argument graphs from natural language texts.

%
%
%
\bibliographystyle{splncs04}
\bibliography{reference}

\end{document}